\documentclass[preprint]{article}
\usepackage{microtype}
\usepackage{graphicx}
\usepackage{subfigure}
\usepackage{booktabs} % for professional tables
\usepackage{soul} % for \ul (underline) command
\usepackage[table]{xcolor} % for \cellcolor
\usepackage{multirow} % for \multirow command
\usepackage{tabularx} % for tabularx environment

\usepackage{hyperref}
\usepackage{float}

\usepackage{icml2025}

\usepackage{amsmath}
\usepackage{amssymb}
\usepackage{mathtools}
\usepackage{amsthm}

\usepackage[capitalize,noabbrev]{cleveref}

\theoremstyle{plain}

\theoremstyle{definition}

\theoremstyle{remark}

\usepackage[dvipsnames]{xcolor}
\usepackage[most]{tcolorbox}
\usepackage{listings}
\usepackage{enumitem}
\newtcolorbox{dialogbox2}[1][]{
  arc=4mm,
  colback=lightgray!20,
  colframe=green!25!black,
  rounded corners,
  boxrule=1.5pt,
  fonttitle=\sffamily\bfseries,
  coltitle=white,
  toptitle=2mm,
  bottomtitle=2mm,
  title=#1, 
  frame style={dashed}, 
  breakable, 
}
\newtcolorbox{dialogbox}[1][]{
  arc=4mm,
  colback=lightgray!20,
  colframe=green!30!black,
  rounded corners,
  boxrule=1.5pt,
  fonttitle=\sffamily\bfseries,
  coltitle=white,
  toptitle=1mm,
  bottomtitle=1mm,
  title=#1, 
  frame style={dashed}, 
  breakable,  
}
\usepackage[textsize=tiny]{todonotes}

\icmltitlerunning{Beyond Imitation: Reinforcement Learning for Active Latent Planning}

\begin{document}

\twocolumn[
\icmltitle{Beyond Imitation: Reinforcement Learning for Active Latent Planning}




\begin{icmlauthorlist}
\icmlauthor{Zhi Zheng}{nus}
\icmlauthor{Wee Sun Lee}{nus}
\end{icmlauthorlist}

\icmlaffiliation{nus}{School of Computing, National University of Singapore, Singapore}

\icmlcorrespondingauthor{Wee Sun Lee}{leews@comp.nus.edu.sg}

\icmlkeywords{Machine Learning, ICML}

\vskip 0.3in
]



\printAffiliationsAndNotice{}  % leave blank if no need to mention equal contribution

\begin{abstract}
Aiming at efficient and dense chain-of-thought (CoT) reasoning, latent reasoning methods fine-tune Large Language Models (LLMs) to substitute discrete language tokens with continuous latent tokens. These methods consume fewer tokens compared to the conventional language CoT reasoning and have the potential to plan in a dense latent space. However, current latent tokens are generally supervised based on imitating language labels. Considering that there can be multiple equivalent but diverse CoT labels for a question, passively imitating an arbitrary one may lead to inferior latent token representations and latent reasoning policies, undermining the potential planning ability and resulting in clear gaps between training and testing. In this work, we emphasize the importance of active planning over the representation space of latent tokens in achieving the optimal latent reasoning policy. So, we propose the \underline{A}c\underline{t}ive Latent \underline{P}lanning method (ATP-Latent), which models the supervision process of latent tokens as a conditional variational auto-encoder (VAE) to obtain a smoother latent space. Moreover, to facilitate the most reasonable latent reasoning policy, ATP-Latent conducts reinforcement learning (RL) with an auxiliary coherence reward, which is calculated based on the consistency between VAE-decoded contents of latent tokens, enabling a guided RL process. In experiments on LLaMA-1B, ATP-Latent demonstrates +4.1\% accuracy and -3.3\% tokens on four benchmarks compared to advanced baselines. Codes are available on \url{https://github.com/zz1358m/ATP-Latent-master}.
\end{abstract}

\begin{figure}[t]
    \centering
    \subfigure[Equivalent Language CoTs for a question]{\includegraphics[width = 0.46\textwidth]{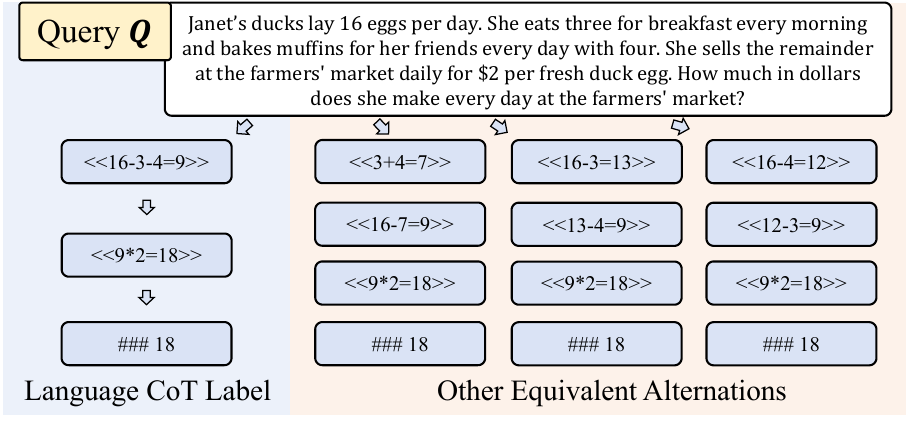}}
    \subfigure[Existing Imitation-based Latent Reasoning methods]{\includegraphics[width = 0.46\textwidth]{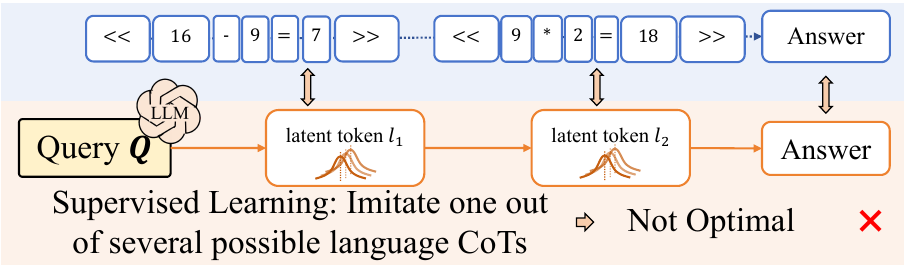}}
    \subfigure[Active Planning for better latent reasoning policies (Ours)]{\includegraphics[width = 0.46\textwidth]{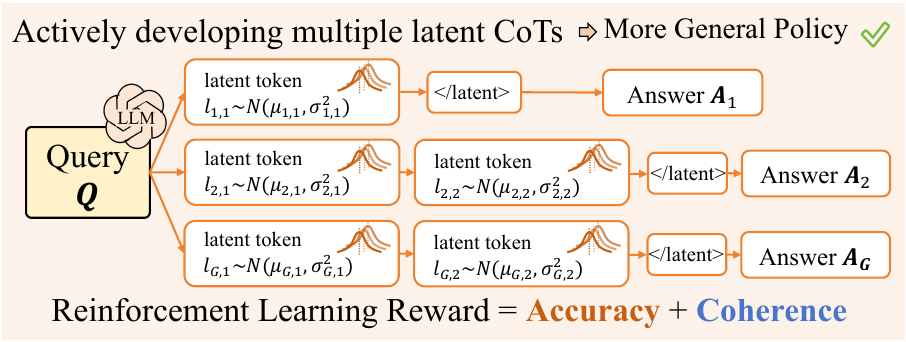}}\vspace{-5pt}
    \caption{Equivalent Language CoTs may lead to different latent reasoning policies. Existing latent reasoning methods (b) imitate one of them, leading to suboptimal policies. The proposed ATP-Latent method (c) actively optimizes the latent reasoning policies in a well-defined space, employing both the verifiable accuracy of answers and the coherence of decoded latent CoTs as rewards.}\label{fig:figure1}\vspace{-10pt}
\end{figure}

\section{Introduction}
Large Language models (LLMs) have shown remarkable capabilities in mathematical reasoning \cite{li2025system}, especially equipped with the chain-of-thought (CoT) prompting techniques \cite{wei2022chain}. In pursuit of superior reasoning ability, LLMs can be further refined with supervised fine-tuning (SFT) \cite{hu2022lora,xia2025tokenskip} or reinforcement learning (RL) techniques \cite{shao2024deepseekmath}. However, although significant improvements in reasoning ability are achieved, these techniques usually lead to significantly longer CoT reasoning paths \cite{liu2025efficient,feng2025efficient}, which will hinder the application of LLMs to scenarios that require real-time responses.

To reduce the latency of LLM-based reasoning, recent work has explored \textbf{implicit latent reasoning}, which aims to enable LLMs to represent and reason in a continuous latent space \cite{hao2024training, tan2025think, shen2025codi, wei2025sim, deng2025latent}. As shown in Figure~\ref{fig:figure1}(b), latent reasoning methods feed high-dimensional embeddings (named latent tokens)---rather than sampled discrete language tokens---into the model at the next LLM step. These methods can avoid redundant natural-language tokens (e.g., those used mainly for semantic cohesion or interpretability-oriented summarization), thus increasing the information density per token and improving the reasoning performance under a fixed token budget.

In implementing the language-to-latent token shift, existing methods generally adopt an imitation-based strategy, mapping the thinking policy of each part of the language CoTs to a latent token through token-by-token curriculum learning \cite{hao2024training}, self-distillation \cite{shen2025codi}, or compression \cite{tan2025think}. These methods can achieve clear reductions in token cost. However, as shown in Figure \ref{fig:figure1} (a), there are usually several equivalent correct language CoT for each question, so learning the latent reasoning policy by imitating an arbitrary one of them will lead to overfitted latent token representations and biased latent policies, resulting in clear performance gaps between training and testing.

To achieve more generalizable latent reasoning, this paper highlights the importance of active planning over latent tokens in pursuing the optimal latent reasoning policy. To this end, as shown in Figure~\ref{fig:figure1}(c), we propose the two-stage \textbf{\underline{A}c\underline{t}ive Latent \underline{P}lanning} (ATP-Latent) method. In the SFT-stage, ATP-Latent extends the original imitation-based methods to train a variational auto-encoder \citep{kingma2013auto} (VAE) for interpretability and a smooth latent token representation space. Unlike prior latent reasoning methods that only passively imitate a single labeled CoT trajectory, ATP-Latent actively optimizes latent reasoning policies in the RL stage, employing both the final answer's correctness (i.e., Accuracy in Figure~\ref{fig:figure1}(c)) and the consistency of VAE-decoded contents (i.e., Coherence in Figure~\ref{fig:figure1}(c)) as rewards, providing soft constraints for the RL state space. We conduct experiments over LLaMA-1B LLMs, demonstrating that ATP-Latent achieves superior reasoning efficiency and accuracy with fewer latent tokens compared with prior SFT and imitation-only latent reasoning methods. Our main contributions are summarized as follows:
\begin{itemize}[leftmargin=*] 
    \item ATP-Latent highlights the significance of active reasoning path planning in finding a good latent reasoning policy, introducing VAE and stop-head for a smoother latent token representation space.
    \item In the RL process, the proposed ATP-Latent method utilizes the supervised VAE decoder to decode latent tokens and employs the consistency between the VAE-decoded contents as auxiliary rewards, yielding unsupervised but beneficial RL training signals for latent planning, providing soft constraints for the RL state space.
    \item ATP-Latent with LLaMA-1B demonstrates 4.1\% more accuracy and 3.3\% fewer tokens on numerical reasoning benchmarks compared to advanced baselines.
\end{itemize}

\section{Preliminaries}

\subsection{Language LLM Reasoning} \label{languageCoT}
The language reasoning process addresses a given question $\boldsymbol{Q}=(q_1,\ldots,q_{|\boldsymbol{Q}|})$ by first generating a series of CoT language reasoning tokens $\boldsymbol{R}=(r_1,\ldots,r_{|\boldsymbol{R}|})$, followed by answer tokens $\boldsymbol{A}=(a_1,\ldots,a_{|\boldsymbol{A}|})$. All tokens involved exist in the discrete language domain, meaning $\boldsymbol{Q}, \boldsymbol{R}, \boldsymbol{A} \subset \mathcal{T}^*$, where $\mathcal{T}$ denotes the full set of language tokens. Both reasoning and answer tokens are produced according to the next-token prediction policy $\pi_\theta$ of large language models (LLMs), as modeled by:
\begin{equation}
\begin{aligned}
p(\boldsymbol{R},\boldsymbol{A}|\boldsymbol{Q}) = &\prod_{t=1}^{|\boldsymbol{R}|}\pi_{\theta}(r_t|[\boldsymbol{Q},\boldsymbol{r}_{1:t-1}])\\
& \qquad \prod_{t=1}^{|\boldsymbol{A}|}\pi_{\theta}(a_t|[\boldsymbol{Q},\boldsymbol{R},\boldsymbol{a}_{1:t-1}]),
\end{aligned}\label{problanguage}
\end{equation}
where $\boldsymbol{r}_{1:t-1}=(r_1,\ldots,r_{t-1})$ and $\boldsymbol{a}_{1:t-1}=(a_1,\ldots,a_{t-1})$; $[\cdot,\cdot]$ as well as $[\cdot,\cdot,\cdot]$ denote concatenation.

\begin{figure*}[htbp]\vspace{-5pt}
\begin{equation}
\begin{aligned}
&\mathcal{J}_{\text{GRPO}}(\theta) = \frac{1}{G} \mathbb{E}_{\boldsymbol{Q}\sim \mathcal{D}, \{\hat{\boldsymbol{L}}\}_{g=1}^G, \{\boldsymbol{A}\}_{g=1}^G \sim p(\cdot,\cdot|\boldsymbol{Q})}\Bigg[\sum_{g=1}^G \frac{1}{\left| \hat{\boldsymbol{L}}_g \right|+\left| \boldsymbol{A}_g \right|} \sum_{t=1}^{\left| \hat{\boldsymbol{L}}_g \right|+\left| \boldsymbol{A}_g \right|} \Big( \min \left( p_{g,t} \hat{A}_{g}, \text{clip}(p_{g,t}, 1 - \epsilon, 1 + \epsilon)\hat{A}_{g} \right) \Bigg]\\
&\qquad\qquad\hat{A}_{g}=\frac{f(\boldsymbol{A}_{g})-\text{mean}(f(\boldsymbol{A}))_{g=1}^G}{\text{std}(f(\boldsymbol{A}))_{g=1}^G},\qquad\qquad p_{g,t} = 
\begin{cases} 
\frac{\pi_{\theta}(a_{g,t} | [\boldsymbol{Q}, \hat{\boldsymbol{L}}_g, (a_{g,1}, \ldots, a_{g,t-1})])}{\pi_{\theta_{\text{old}}}(a_{g,t} | [\boldsymbol{Q}, \hat{\boldsymbol{L}}, (a_{g,1}, \ldots, a_{g,t-1})])} & \text{if } t > |\hat{\boldsymbol{L}}_g| \\[5pt]
\frac{\pi_{\theta}(\hat{\boldsymbol{l}}_{g,t} | [\boldsymbol{Q}, (\hat{\boldsymbol{l}}_{g,1}, \ldots, \hat{\boldsymbol{l}}_{g,t-1})])}{\pi_{\theta_{\text{old}}}(\hat{\boldsymbol{l}}_{g,t} | [\boldsymbol{Q}, (\hat{\boldsymbol{l}}_{g,1}, \ldots, \hat{\boldsymbol{l}}_{g,t-1})])} & \text{if } t \leq |\hat{\boldsymbol{L}}_g|.
\end{cases}
\end{aligned}\label{grpo}
\end{equation}
\end{figure*}

\paragraph{SFT for Language LLM Reasoning} In SFT approaches for language-based LLM reasoning, carefully annotated CoT sequences $[\boldsymbol{R}^*,\boldsymbol{A}^*]$ are collected for each question $\boldsymbol{Q}$. The LLM is then fine-tuned to maximize accuracy over all $|\boldsymbol{R}^*| + |\boldsymbol{A}^*|$ token positions. The effectiveness of SFT is highly dependent on the quality of the CoT labels, which are challenging to obtain for complex reasoning datasets.

\paragraph{RL Fine-tuning for Language LLM Reasoning} RL methods—such as Group Relative Policy Optimization (GRPO) \cite{liu2024deepseek}, Dr. GRPO \cite{liu2025understanding}, DAPO \cite{yu2025dapo}, and Lite PPO \cite{liu2025part}—sample multiple candidate CoTs $[\boldsymbol{R}, \boldsymbol{A}]$ per question and assess each with a reward reflecting the answer quality $\boldsymbol{A}$. For example, in standard GRPO \cite{shao2024deepseekmath}, $G$ candidate CoTs are generated for each $\boldsymbol{Q}$, and the objective is updated according to the relative advantage within these $G$ samples. These RL-based approaches often surpass SFT in mathematical reasoning tasks. However, they have a tendency to improve performance by progressively generating longer CoT trajectories over time \cite{liu2025understanding, liu2025part}, a phenomenon termed "overthinking" \cite{sui2025stop}, which can significantly reduce the inference efficiency of reasoning LLMs.

% \textbf{Fine-tuning for Concise Language CoT.} There are methods focusing on generating concise language CoTs without severely losing effectiveness \cite{feng2025efficient, xu2025scalable}. Some of these methods seek to heuristically detect and remove unnecessary tokens from the CoT label $[\boldsymbol{R}^*,\boldsymbol{A}^*]$ for SFT \cite{su2025token,qiao2025concise}. Others formulate the fine-tuning process for concise language CoT as a multi-objective optimization task \cite{qi2025optimizing} and modify the reward function of the RL process (e.g., $f(\boldsymbol{A}_{t})=1-\alpha|\boldsymbol{A}_{t}|$) to improve the overall performance within any thinking budgets \cite{arora2025training,dai2025s}.

\begin{figure*}[htbp]
    \centering
    \subfigure[SFT stage of ATP-Latent]{\includegraphics[width = 0.52\textwidth]{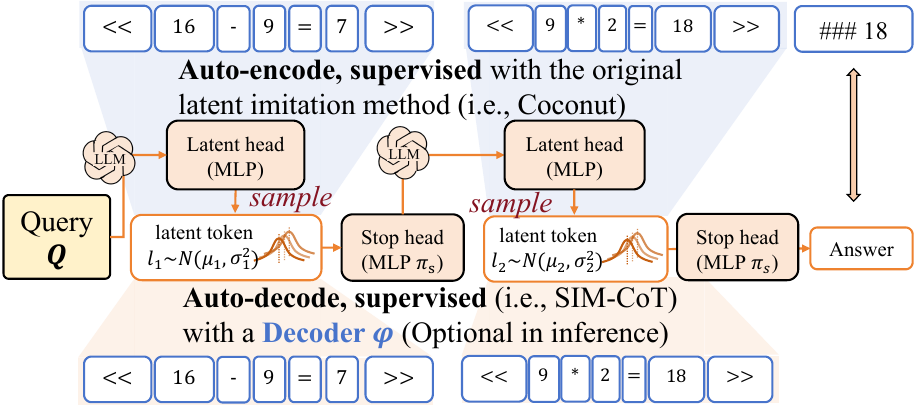}}
    \subfigure[RL stage of ATP-Latent]{\includegraphics[width = 0.475\textwidth]{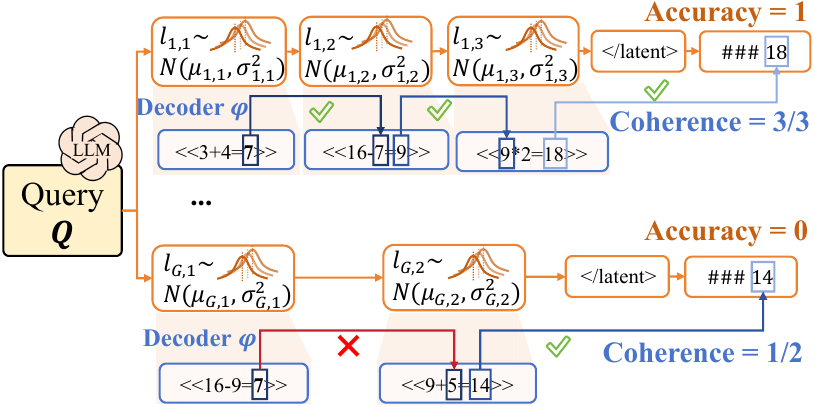}}
    \caption{Equivalent Language CoTs represent different reasoning policies. Existing latent reasoning methods (b) imitate one out of them, leading to suboptimal latent policies. The proposed ATP-Latent method (c) actively optimizes latent policies, employing both the accuracy of answers and the coherence of latent CoTs as reward.}\label{fig:figure2}
\end{figure*}

\subsection{Latent CoT reasoning} 
To achieve efficient reasoning, recent works have attempted to stimulate concise language reasoning \cite{feng2025efficient,liu2025efficient,xia2025tokenskip}. As another more effective but difficult way to solve overthinking, latent CoT reasoning methods \cite{chen2025reasoning} aim at shifting the original reasoning process in the pre-trained language domain to $|\boldsymbol{L}|$ continuous latent tokens $\boldsymbol{L}=(\boldsymbol{l}_1,\ldots,\boldsymbol{l}_{|\boldsymbol{L}|})$. Every latent token $\boldsymbol{l}_i\in \mathcal{R}^{d},\ i\in\{1,\ldots,|\boldsymbol{L}|\}$ is a $d$-dimensional real vector. The latent reasoning LLMs still generate the answer part $\boldsymbol{A}$ in natural language after $\boldsymbol{L}$. In generating latent tokens, most latent reasoning methods follow the deterministic next(-latent)-token-prediction paradigm as follows:
\begin{equation}
\begin{aligned}
\forall\ t\leq |\boldsymbol{L}|,\quad \boldsymbol{l}_t = \pi_{\theta}([\boldsymbol{Q},\boldsymbol{l}_{1:t-1}]),
\end{aligned}\label{deterlatent}
\end{equation}
where $\boldsymbol{l}_{1:t-1}=(\boldsymbol{l}_1,\ldots,\boldsymbol{l}_{t-1})$. As shown in Figure \ref{fig:figure1}(b), most existing approaches to latent CoT reasoning fine-tune pre-trained LLMs to imitate language labels, effectively mapping latent tokens to mimic the role of one or more language tokens \cite{wang2025synadapt, cheng2024compressed, shen2025codi}. \citet{hao2024training} first enables language-to-latent domain transformation by progressively substituting original tokens in $\boldsymbol{R}$ with latent tokens, following a curriculum-based strategy. In the $k$-th curriculum stage, it substitutes language tokens in the first $k$ steps with $c*k$ latent tokens $\boldsymbol{l}_1,\ldots,\boldsymbol{l}_{c*k}$ ($c$ latent tokens for each step) by minimizing the loss of cross-entropy (CE) to predict the rest of the part. This process fits well with the purpose of predicting the next-latent-token. Besides, \citet{shen2025codi} fine-tunes LLM via self-distillation to encourage several latent tokens imitating the hidden state of the final token, \citet{tan2025think} and \citet{deng2025latent} directly compress the embeddings of several consensus language tokens, \citet{wei2025sim} introduces supervision on latent tokens, training another LLM to explain latent tokens back to language steps it imitates. 

These methods offer a clear reduction in token cost. As another possible advantage on effectiveness \cite{hao2024training}, reasoning in the latent token space can effectively eliminate some redundant tokens that are purely for syntax or explanation, thus implementing dense CoT reasoning. However, as shown in Figure \ref{fig:figure1}(a), there may be several correct CoTs for one question, so imitating the implicit reasoning policy from an arbitrary label will lead to inferior latent reasoning policies and token representations, demonstrating a clear gap between training and testing. So, to find the most general latent reasoning policy, this paper proposes to highlight the planning ability of latent reasoning LLMs after active latent reasoning path exploration via RL. 

\section{Studies on Exploring Latent CoTs}\label{section3}

As the main obstacle to latent reasoning exploration, current latent reasoning methods usually predict a deterministic latent token $\boldsymbol{l}_t$ at each step. So, in reinforcing latent policies with exploration, recent works \citet{butt2025soft,tan2025think} introduce Gaussian Noise as $\hat{\boldsymbol{l}}_t =\boldsymbol{l}_t+\mathcal{N}(0,\sigma^2 Id)$ on latent tokens and do GRPO over the latent tokens with Gaussian reparameterization. The GRPO objective of these methods is shown in Eq. \eqref{grpo}, where $\hat{A}_{g}$ represents the advantage function for the $g$-th latent CoT $\hat{\boldsymbol{L}}_g=(\hat{\boldsymbol{l}}_{g,1}, \ldots, \hat{\boldsymbol{l}}_{g,|\hat{\boldsymbol{L}}_g|})$, the reward function $f(\boldsymbol{A}_{g})=1$ if and only if the answer $\boldsymbol{A}_{g}$ is correct, $\epsilon$ is the clipping hyperparameter, $\pi_{\theta_\text{old}}$ is the sampling policy before updates, and we follow \citet{yu2025dapo} and \citet{tan2025think} in removing the KL term to facilitate new latent policies besides the one from SFT. The reparameterization probability is calculated as follows:
\begin{equation}
\begin{aligned}
    &\pi_{\theta}(\hat{\boldsymbol{l}}_{t} | [\boldsymbol{Q}, \boldsymbol{l}_{1:t-1}]) \propto \exp\left(-\frac{1}{2\sigma^2} \|\hat{\boldsymbol{l}}_t - \boldsymbol{l}_t\|^2_2\right),\\
    & \qquad \text{where} \quad \boldsymbol{l}_t = \pi_{\theta}([\boldsymbol{Q},(\hat{\boldsymbol{l}}_1,\ldots,\hat{\boldsymbol{l}}_{t-1})]).
\end{aligned}
\end{equation}
However, these methods do not demonstrate clear improvements in accuracy in primary settings \citep{tan2025think}. This might be due to their failure to maintain a smooth state space (i.e., the latent token representation space) for RL. Moreover, these methods rely on random exploration over the latent space without any confines. Both drawbacks make it hard for RL to explore a good latent reasoning policy.
\begin{figure}[h]
    \centering
    \includegraphics[width=0.8\linewidth]{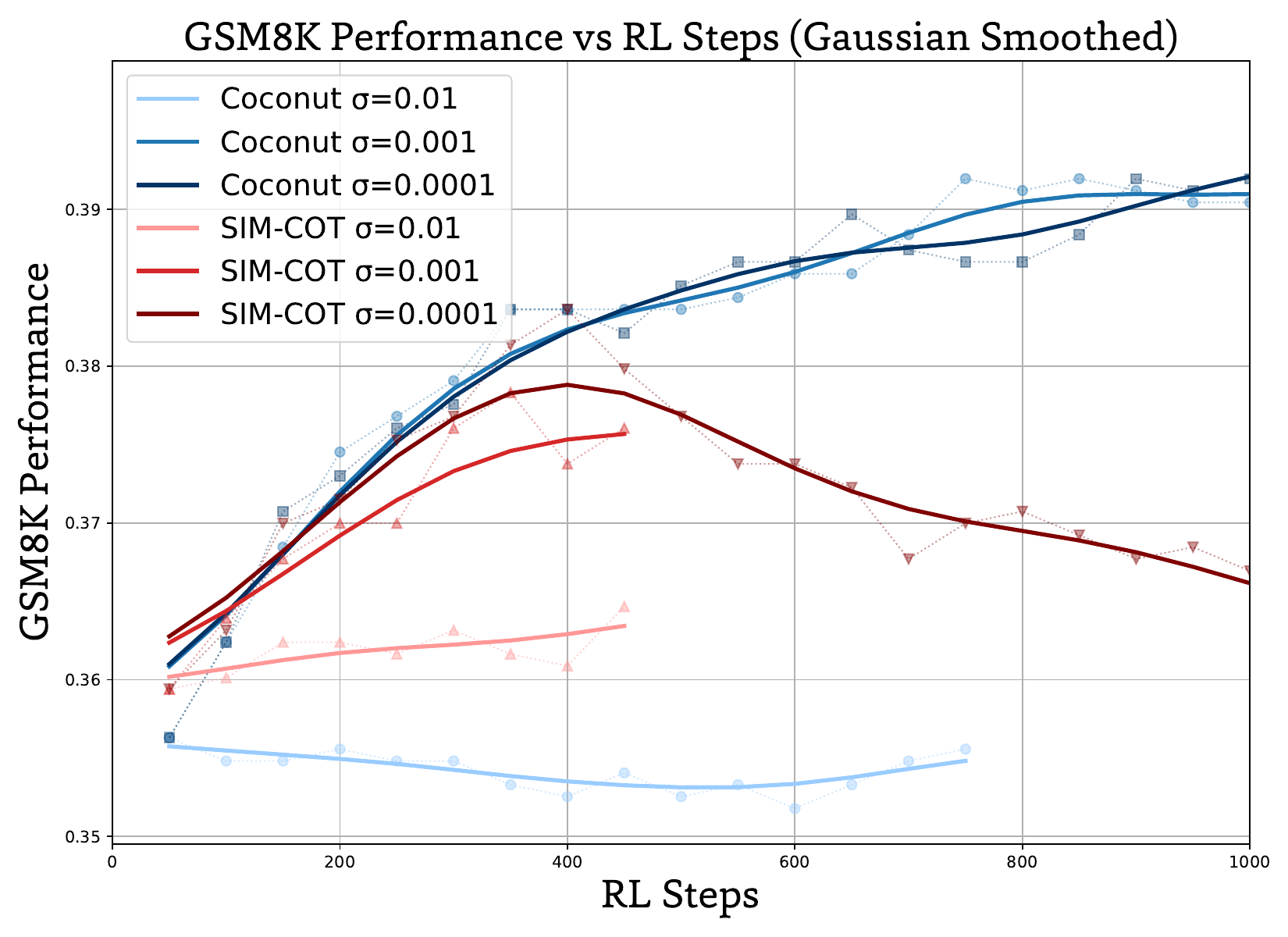}
    \caption{GRPO validation curve over Coconut and SIM-CoT and noises in different scales.}\vspace{-12pt}
    \label{fig:Coconut}
\end{figure}

\begin{figure}[htbp]
    \centering
    \includegraphics[width=0.8\linewidth]{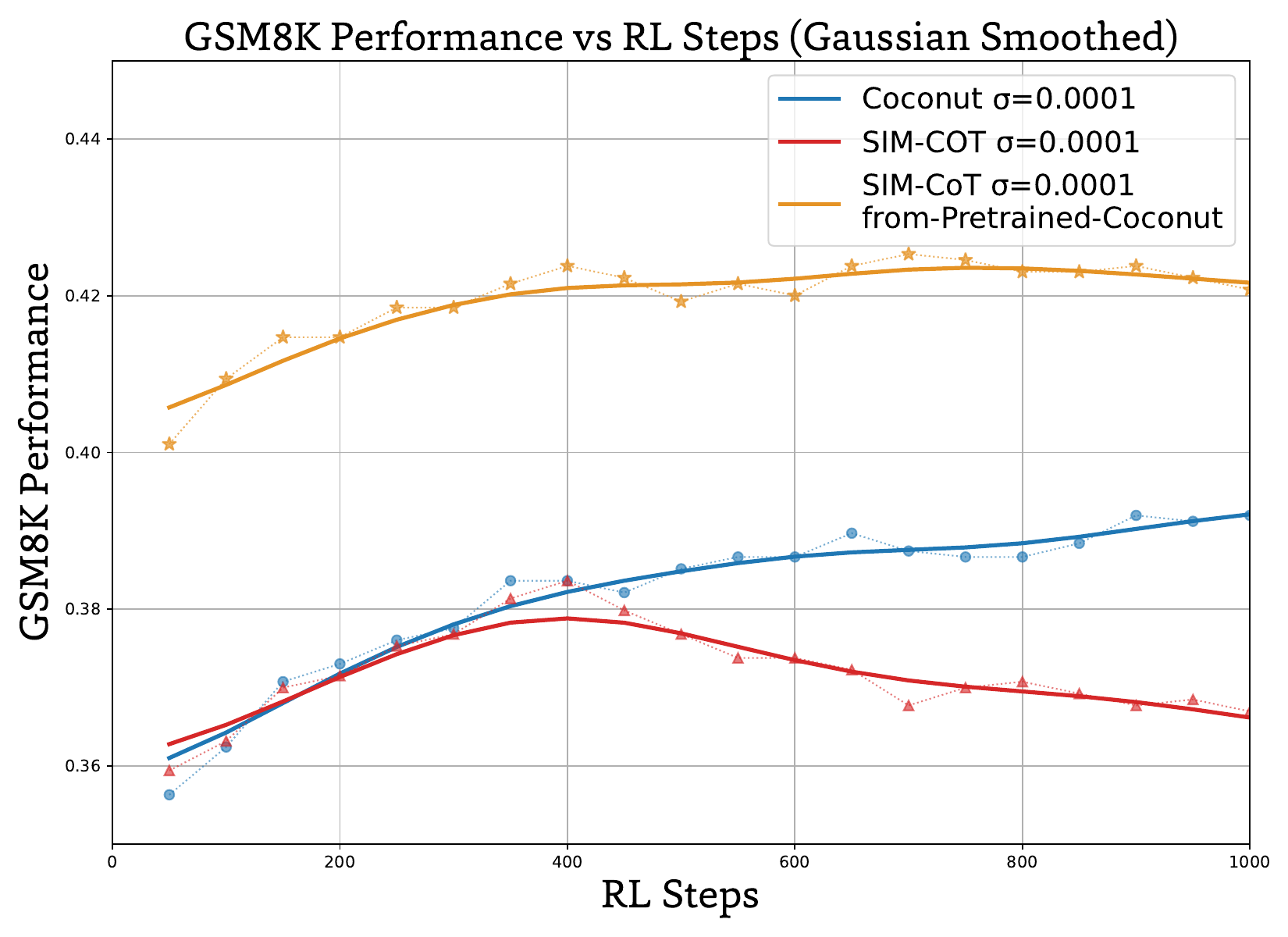}
    \caption{GRPO validation curve of SIM-CoT finetuned after Coconut training.}
    \label{fig:simcot}
    \vspace{-12pt}
\end{figure}

In demonstrating this, as shown in Figure \ref{fig:Coconut}, we implement the standard GRPO in Eq. \ref{grpo} over Coconut \cite{hao2024training} and its upgrade SIM-CoT \cite{wei2025sim}, which introduces a decoder to recover the imitation. Surprisingly, with appropriate $\sigma$ settings, Coconut can achieve a significant improvement through simple reparameterization and RL. However, its upgraded version, SIM-CoT, despite offering better latent space supervision and much better benchmark performances, fails to show stable benefits with RL. However, as shown in Figure \ref{fig:simcot}, running SIM-CoT on a pre-trained Coconut before the RL process activates its exploration capabilities. Therefore, we attribute this difference to the SIM-CoT sharpening of the latent token representation space, thus limiting the exploration capabilities.

So, to provide a \textbf{smooth} latent token representation space conducive to RL exploration, and to leverage the explainability introduced in SIM-CoT \citep{wei2025sim} for a well-defined RL action space, we propose the ATP-Latent method. Specifically, ATP-Latent: \textbf{1)} Models the latent imitation process by training a VAE \citep{higgins2017beta} over language CoT labels, promoting uniformity across the representation space of latent tokens; \textbf{2)} Employs an automatic stopping strategy during latent token generation, ensuring that each latent token encodes a comparable amount of information and thereby unifies the latent space; \textbf{3)} Utilizes the VAE decoder to decode the latent tokens in the RL stage and introduces an unsupervised coherence reward to judge the consistency of generated contents, thus providing soft constraints on the RL state space.

\section{Method: Active Latent Planning}

In this section, we propose the Active Latent Planning (ATP-Latent) method (shown in Figure \ref{fig:figure2}). It contains an SFT stage and a follow-up RL stage. In the SFT stage, ATP-Latent models the imitation process of the language label by \textbf{VAE-like training}, taking the LLM reasoning pipeline as the encoder and an additional LLM as the decoder for revealing each latent token to languages. We also enable the \textbf{stop mechanism} for smooth latent representations. In the following RL stage, ATP-Latent not only takes the accuracy of answers as a reward, but also measures the \textbf{coherence of the latent tokens} by the consistency between VAE-decoded contents, providing soft constraints on the RL state space.

\subsection{SFT Stage}

The SFT stage of ATP-Latent aims to maintain a smooth latent token representation space for the subsequent RL stage. To achieve this goal, we extend the latent imitation with the explainer idea in SIM-CoT \citep{wei2025sim} to imitate language tokens as training VAEs. 

\paragraph{Encoder} ATP-Latent adopts the original imitation-based latent reasoning model (with parameter $\theta$) as the VAE encoder, and following \citet{wei2025sim}, we use another LLM (with parameter $\phi$) as the VAE decoder. When using the Coconut \citep{hao2024training} method as the ATP-Latent imitation process, in the $k$-th curriculum stage, it removes the first $k$ steps (noted $r_1$ to $r_{n(k)}$) in the CoT language label $\boldsymbol{R}$, generates $c*k$ latent tokens $\boldsymbol{l}_1,\ldots,\boldsymbol{l}_{c*k}$ in latent token representation space by, and minimizes the CE loss of the rest part ($r_{n(k)+1}$ to $r_{|\boldsymbol{R}|}$) as follows:
\begin{equation}
\begin{aligned}
&\mathcal{L}_{\text{Enc}} = - \sum_{t=n(k)+1}^{|\boldsymbol{R}|} \log \pi_\theta (r_t |[\boldsymbol{Q}, \boldsymbol{l}_{1:c*k}, \boldsymbol{r}_{n(k)+1:t-1}]) \\&\quad -\sum_{t=1}^{|\boldsymbol{A}|} \log \pi_\theta (a_t |[\boldsymbol{Q}, \boldsymbol{l}_{1:c*k}, \boldsymbol{r}_{n(k)+1:|\boldsymbol{R}|}, \boldsymbol{a}_{1:t-1}]).
\end{aligned}
\end{equation}
The original Coconut and SIM-CoT generate deterministic latent tokens $\boldsymbol{L}$, which may cause the latent token representation to overfit the label, undermining the expression of situations in Figure \ref{fig:figure1} where there may be multiple equivalent latent policies, thus hindering the subsequent RL finetuning. So, in implementing the Gaussian distributions for VAE, we add a Multi-Layer Perceptron (MLP) as the latent head over the output hidden state, predicting the mean $\boldsymbol{\mu_t}\in \mathcal{R}^{d}$ and standard deviation $\boldsymbol{\sigma_t}\in \mathcal{R}^{d}$ by the latent reasoning LLM with parameter $\theta$ as follows:
\begin{equation}
\forall\ t\leq |\boldsymbol{L}|,\quad \boldsymbol{\mu_t},\boldsymbol{\sigma_t} =  \theta([\boldsymbol{Q},(\boldsymbol{l}_1,\ldots,\boldsymbol{l}_{t-1})]).
\end{equation}
Then, latent tokens are sampled accordingly as follows:
\begin{equation}
\forall\ t\leq |\boldsymbol{L}|,\quad \boldsymbol{l}_t \sim \mathcal{N}(\boldsymbol{\mu_t},\mathrm{diag}(\boldsymbol{\sigma}_t^2)).
\end{equation}
\paragraph{Stop Head of Encoder} Most latent reasoning methods choose to generate fixed-length latent tokens during the SFT stage, which may lead to inconsistencies in the information density between latent tokens. To achieve a more uniform and smooth latent token representation space for the following planning stage, ATP-Latent designs to implement the stop policy in the CoT label $\boldsymbol{R}$ with $K$ steps. We add one MLP $\pi_{\theta_{s}}$ as the stop head on each sampled latent token $\boldsymbol{l}_t$ for $\pi_{\theta_{s}}(\text{stop}|\boldsymbol{l}_t)$ and $\pi_{\theta_{s}}(\text{cont.}|\boldsymbol{l}_t)$, training the head only to choose stop after the $K$ step with a Binary CE (BCE) loss $\mathcal{L}_\text{Stop}$ as follows:
\begin{equation}
\begin{aligned}
    \mathcal{L}_\text{Stop} = -\sum_{t=1}^{\min(K,c*k)-1}& \log \pi_{\theta_{s}}(\text{cont.}|\boldsymbol{l}_t)\\ -&\sum_{t=\min(K,c*k)}^{c*k} \log \pi_{\theta_{s}}(\text{stop}|\boldsymbol{l}_t).    
\end{aligned}
\end{equation}

\paragraph{Decoder} For each of the $k$ stages. The decoder takes the series of $c$ generated latent tokens as input, aiming at recovering the original language label $\boldsymbol{r}_{n(t-1)+1:n(t)}=(r_{n(t-1)+1},\ldots,r_{n(t)})$. It optimizes the CE Loss as follows:
\begin{equation}
\begin{aligned}
    \mathcal{L}_{\text{Dec}} =& -\sum_{t=1}^{k}\sum_{m=n(t-1)+1}^{n(t)} \\&\log \pi_{\phi}\big(r_m|[\boldsymbol{l}_{(t-1)*c+1:t*c},\boldsymbol{r}_{n(t-1)+1:m-1}]\big).
\end{aligned}
\end{equation}
When $k>K$, there is no content to decode, so the decoder needs to output an empty string. As the standard setting in the VAE training process, we also optimize the KL loss based on the ELBO \citep{kingma2013auto} as follows:
\begin{equation}
\begin{aligned}
\mathcal{L}_\text{KL} &= D_{\mathrm{KL}}\left(
\mathcal{N}(\boldsymbol{\mu_t},\, \boldsymbol{\sigma_t}^2Id)) \ \Vert\ 
\mathcal{N}(0,\, I)\right)\\ 
&= \frac{1}{2} \sum_{t=1}^{c*k}\sum_{j=1}^d
\left(
\mu_{t,j}^2 + \sigma_{t,j}^2 - 1 - \log \sigma_{t,j}^2
\right).
\end{aligned}
\end{equation}
Finally, the loss function in the SFT stage is designed with the $\beta$-VAE technique \citep{higgins2017beta} as follows:
\begin{equation}
    \mathcal{L}_\text{SFT} = \mathcal{L}_{\text{Enc}} + \mathcal{L}_{\text{Dec}} + \mathcal{L}_\text{Stop} + \beta \mathcal{L}_\text{KL}.
\end{equation}
For $\mathcal{L}_{\text{Dec}}$, $\mathcal{L}_\text{Stop}$ and $\mathcal{L}_{\text{KL}}$, we additionally normalize them by dividing the number of stages $k$.

The latent space often represents infinitely different latent tokens (representing infinite categories of reasoning steps).  Moreover, as shown in Figure \ref{fig:Coconut}, each latent token tends to have only a small range of mutable values (far different from the prior $\mathcal{N}(0,\, I)$). So, taking into account these properties, we set a very small $\beta=\frac{1e-3}{d}$ to prioritize the latent reasoning process over decoding.

\begin{table*}[t]
\centering
\caption{
Comparison of (latent) reasoning methods and their ablations on four math reasoning benchmarks (GSM8K, GAM-hard, MultiArith, SVAMP). We report the answer accuracy (noted Acc, higher is better) and the average number of generated tokens $|\boldsymbol{L}|+|\boldsymbol{A}|$ (\#Token, lower is better) for each approach. We bolded the best-performing method for each benchmark and highlighted cells indicate the second-best performance in each setting. * represents the results reported in \citet{wei2025sim}. The proposed ATP-Latent and its ablations demonstrate the effectiveness of latent reasoning with the proposed components, including VAE, Stop head, and the RL stage. \textit{w} Only Coherence as Reward represents using only the unsupervised coherence reward in RL training. We show the change of average accuracy in colors. For methods with randomness, we run each method 5 times and report the average. 
}
\renewcommand\arraystretch{1.05}
\setlength{\tabcolsep}{2.3mm}
\resizebox{\textwidth}{!}{
\begin{tabular}{l|cc|cccccc|cc}
\bottomrule[0.5mm]
& \multicolumn{2}{c|}{GSM8K}   & \multicolumn{2}{c}{GAM-hard}& \multicolumn{2}{c}{MultiArith}   & \multicolumn{2}{c|}{SVAMP}   & \multicolumn{2}{c}{Avg} \\ \hline
& Acc & \#Token & Acc& \#Token & Acc& \#Token & Acc & \#Token & Acc & \#Token \\ \hline\hline
CoT-SFT   & \textbf{52.3} & 30.6    & \textbf{12.3}& 35.6    & \cellcolor[HTML]{D9D9D9}\textbf{94.4} & 19.0    & \textbf{58.0} & 16.3    & \textbf{54.3} & 25.4    \\
Answer-SFT& 26.4& 4.2& 6.4& 4.8& 49.4    & 4.0& 35.3& 4.1& 29.4& 4.3\\ \hline
CoLaR \citep{tan2025think}& 25.7& 10.0    & 5.7& 11.5    & 86.8    & 7.2& \cellcolor[HTML]{D9D9D9}49.9 & 7.1& 42.0& 8.9\\
iCoT* \citep{deng2023icot}  & 30.1 & 2.2    & 5.7& -  & 55.5    & -  & 29.4& -  & 30.2& -  \\
Coconut \citep{hao2024training}  & 35.6& 9.2& 8.2& 9.8& 86.1    & 9.0& 37.0& 9.1& 41.7& 9.3\\
Coconut* \citep{hao2024training}  & 36.6& 13.2    & 8.1& -  & 83.5    & -  & 36.2& -  & 41.1& -  \\
SIM-CoT \citep{wei2025sim}   & 35.9& 9.2& 8.6& 9.8& 87.8    & 9.0& 42.0& 9.0& 43.6& 9.2\\
SIM-CoT* \citep{wei2025sim}  & 42.2& 13.2    & 9.3& -  & 87.7    & -  & 43.9& -  & 45.8& -  \\ \hline
ATP-Latent (Ours)& \cellcolor[HTML]{D9D9D9}42.3 & 9.8& \cellcolor[HTML]{D9D9D9}9.8 & 10.0    & \cellcolor[HTML]{D9D9D9}\textbf{94.4} & 7.1& 44.2& 6.8& \cellcolor[HTML]{D9D9D9}47.7 & 8.4\\
-\textit{w/o} VAE   & 40.8& 9.7& 9.2& 10.1    & 92.8    & 7.2& 46.0& 6.4& 47.2{\small\textcolor{Red}{(-0.5)}}    & 8.4\\
-\textit{w/o} Stop Head  & 41.2& 9.2& 8.8& 9.8    & 93.9    & 9.0 & 43.7& 9.0 & 46.9{\small\textcolor{Red}{(-0.8)}}    & 9.3\\
-\textit{w/o} RL (Ours-SFT)   & 40.0& 9.5& 9.3& 9.7& 90.0    & 7.1& 43.7& 6.5& 45.8{\small\textcolor{Red}{(-1.9)}}    & 8.2\\
\ $\cdot$ \textit{w} Only Coherence as Reward & 40.9& 9.4& 9.5& 9.7& 90.6    & 7.1& 46.0& 6.5& 46.7{\small\textcolor{Green}{(+0.9)}}    & 8.2\\ \toprule[0.5mm]
\end{tabular}
}
\label{main}
\end{table*}

\subsection{RL Stage}

The RL stage of ATP-Latent actively explores the representation space of each latent token trained by VAE for the best policy across another training dataset. We design a coherence indicator to provide well-defined guidance for the state space. The RL process is built based on the Eq. \ref{grpo}, where we introduce the stop policy $\pi_{\theta_s}$ to generate $|\boldsymbol{L}_g|$ latent tokens for each sample. We calculate the forward policy $p_t$ probability with $\pi_{\theta_{s}}(\text{cont.}|\boldsymbol{l}_t)$ within the latent CoT and with $\pi_{\theta_{s}}(\text{stop}|\boldsymbol{l}_t)$ at the final latent token as follows:
\begin{equation}
p_{g,t} =
\begin{cases}
\displaystyle
\frac{
    \pi_{\theta}(a_{g,t} | [\boldsymbol{Q}_g, \boldsymbol{L}_g, \boldsymbol{a}_{g,1:t-1}])
}{
    \pi_{\theta_{\text{old}}}(a_{t} | [\boldsymbol{Q}_g, \boldsymbol{L}_g, \boldsymbol{a}_{g,1:t-1}])
}
& t > |\boldsymbol{L}_g| \\[15pt]
\displaystyle
\frac{
    \pi_{\theta_{s}}(\text{stop}|\boldsymbol{l}_{g,t}) 
    \pi_{\theta}(\boldsymbol{l}_{g,t} | [\boldsymbol{Q}_g, \boldsymbol{l}_{g,1:t-1}])
}{
    \pi_{\theta_{s,\text{old}}}(\text{stop}|\boldsymbol{l}_{g,t}) 
    \pi_{\theta_{\text{old}}}(\boldsymbol{l}_{g,t}| [\boldsymbol{Q}, \boldsymbol{l}_{g,1:t-1}])
}
& t = |\boldsymbol{L}_g| \\[15pt]
\displaystyle
\frac{
    \pi_{\theta_{s}}(\text{cont.}|\boldsymbol{l}_{g,t}) 
    \pi_{\theta}(\boldsymbol{l}_{g,t} | [\boldsymbol{Q}, \boldsymbol{l}_{g,1:t-1}])
}{
    \pi_{\theta_{s,\text{old}}}(\text{cont.}|\boldsymbol{l}_{g,t})
    \pi_{\theta_{\text{old}}}(\boldsymbol{l}_{g,t} | [\boldsymbol{Q}, \boldsymbol{l}_{g,1:t-1}])
}
& t < |\boldsymbol{L}_g|.
\end{cases}
\end{equation}
where $\pi_{\theta_{s,\text{old}}}$ is the stop policy in sampling, $\pi_{\theta_{s}}$ represents the current stop policy. $\pi_{\theta_{\text{old}}}$ is the sampling latent policy, $\pi_{\theta}$ is the current latent policy. The probability of $\pi_\theta(\boldsymbol{l}_t|[\boldsymbol{Q},\boldsymbol{l}_{1:t-1}])$ is obtained with the reparameterization trick as follows:
\begin{equation}
    \pi_{\theta}(\boldsymbol{l}_{t}|[\boldsymbol{Q}, \boldsymbol{l}_{1:t-1}]) =
    \prod_{i=1}^d \frac{1}{\sqrt{2\pi \sigma_{t,i}^2}} \exp\left[ -\frac{1}{2} \frac{(l_{t,i}-\mu_{t,i})^2}{\sigma_{t,i}^2 } \right].
\end{equation}
After training the decoder $\phi$ in VAE, it can well translate the latent tokens to languages $\boldsymbol{R}'$ as follows, even with tolerable noise:
\begin{equation}
    \forall_{i=1}^{\lceil\frac{|\mathcal{L}|}{c}\rceil} \quad  \boldsymbol{R}'_i = \pi_{\phi}(\boldsymbol{l}_{(i-1)*c+1},\cdots,\boldsymbol{l}_{i*c}).
\end{equation}
As shown in the \ref{fig:figure2} (b), there are different latent CoTs in the GRPO-based RL sampling, some of which (the lower one) have inconsistent explanations. For math reasoning, ATP-Latent proposes to judge the coherence $R_\text{Coh}(\boldsymbol{L})$ of latent tokens with the consistency of decoded contents, counting the ratio of equations whose \textbf{results} (i.e., right-hand side (RHS)) appear in the \textbf{left-hand side} (LHS) of the following latent CoT steps or the final answer $\boldsymbol{A}$ as follows:
\begin{equation}
\begin{aligned}
R_\text{Coh}(\boldsymbol{L}) &= \frac{\sum_{i=1}^{|\boldsymbol{L}|}\mathbb{I}[\exists_{\boldsymbol{S}\in\mathbb{S}}\ \text{RHS}(\pi_{\phi}(\boldsymbol{L}_{i}))=\boldsymbol{S}]}{\sum_{i=1}^{|\boldsymbol{L}|}1},\\[5pt]
\mathbb{S} &= \text{LHS}(\boldsymbol{R}'_1\cup\cdots\cup\boldsymbol{R}'_{\lceil\frac{|\mathcal{L}|}{c}\rceil}\cup\boldsymbol{A}).
\end{aligned}
\end{equation} 
Generally, we believe latent CoTs with consistent and reasonable explanations can show that they really participate in reasoning, thus representing better policy. Instead, inconsistency between decoded explanations may reflect a wrong latent reasoning policy or even indicate that these latent tokens do not participate in the final answering phase. Existing methods fail to provide reliable rewards over the latent tokens, only being able to explore the general optimal latent policies randomly. ATP-Latent proposes to use the reward $R_\text{Coh}(\boldsymbol{L})$ on latent tokens for soft guidance. So, the reward $f(\boldsymbol{L},\boldsymbol{A})$ of ATP-Latent prioritizes correct answers with latent CoTs in higher $R_\text{Coh}(\boldsymbol{L})$ as follows:
\begin{equation}
\begin{aligned}
&f(\boldsymbol{L},\boldsymbol{A}) =  (1+ 0.1\cdot R_\text{Coh}(\boldsymbol{L}))R_\text{Correct}(\boldsymbol{A}) \\&\qquad \qquad \qquad \qquad \qquad + 0.5\cdot R_\text{Format}(\boldsymbol{A}).
\end{aligned}
\end{equation} 
where $R_\text{Correct}$ and $R_\text{Format}$ represent the binary correctness of the answer and format (please refer to Appendix \ref{prompt}), respectively. It can provide soft constraints on the state space (i.e., latent tokens). Besides, latent reasoning methods sometimes actually build question-to-answer shortcuts instead of reasoning step-by-step \citep{zhang2025latent}, making latent tokens useless. Introducing $R_\text{Coh}(\boldsymbol{L})$ will also be beneficial in avoiding shortcuts.

\paragraph{Coherence as Unsupervised Reward} \label{coh-only} The coherence reward can also serve as a powerful unsupervised reward. So, ATP-Latent provides an unsupervised setting in RL as well, using only the coherence of latent CoTs and format reward rewards as follows:
\begin{equation}
f(\boldsymbol{L},\boldsymbol{A}) = R_\text{Coh}(\boldsymbol{L}) + 0.5\cdot R_\text{Format}(\boldsymbol{A}).
\end{equation} 

\begin{figure*}[htbp]\vspace{-10pt}
    \centering
    \subfigure[RL stage of ATP-Latent]{\includegraphics[width = 0.33\textwidth]{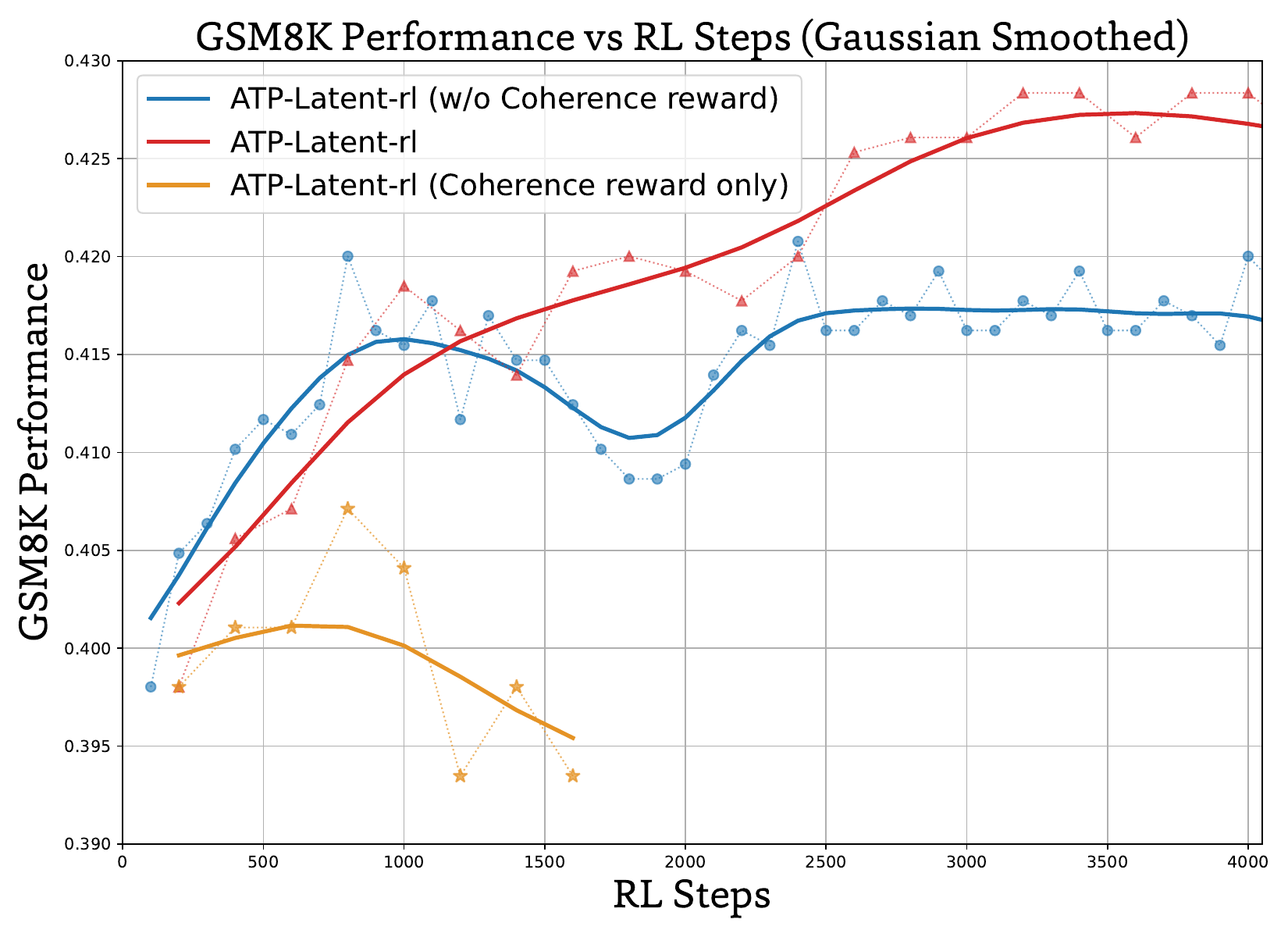}}
    \subfigure[SFT stage of ATP-Latent]{\includegraphics[width = 0.33\textwidth]{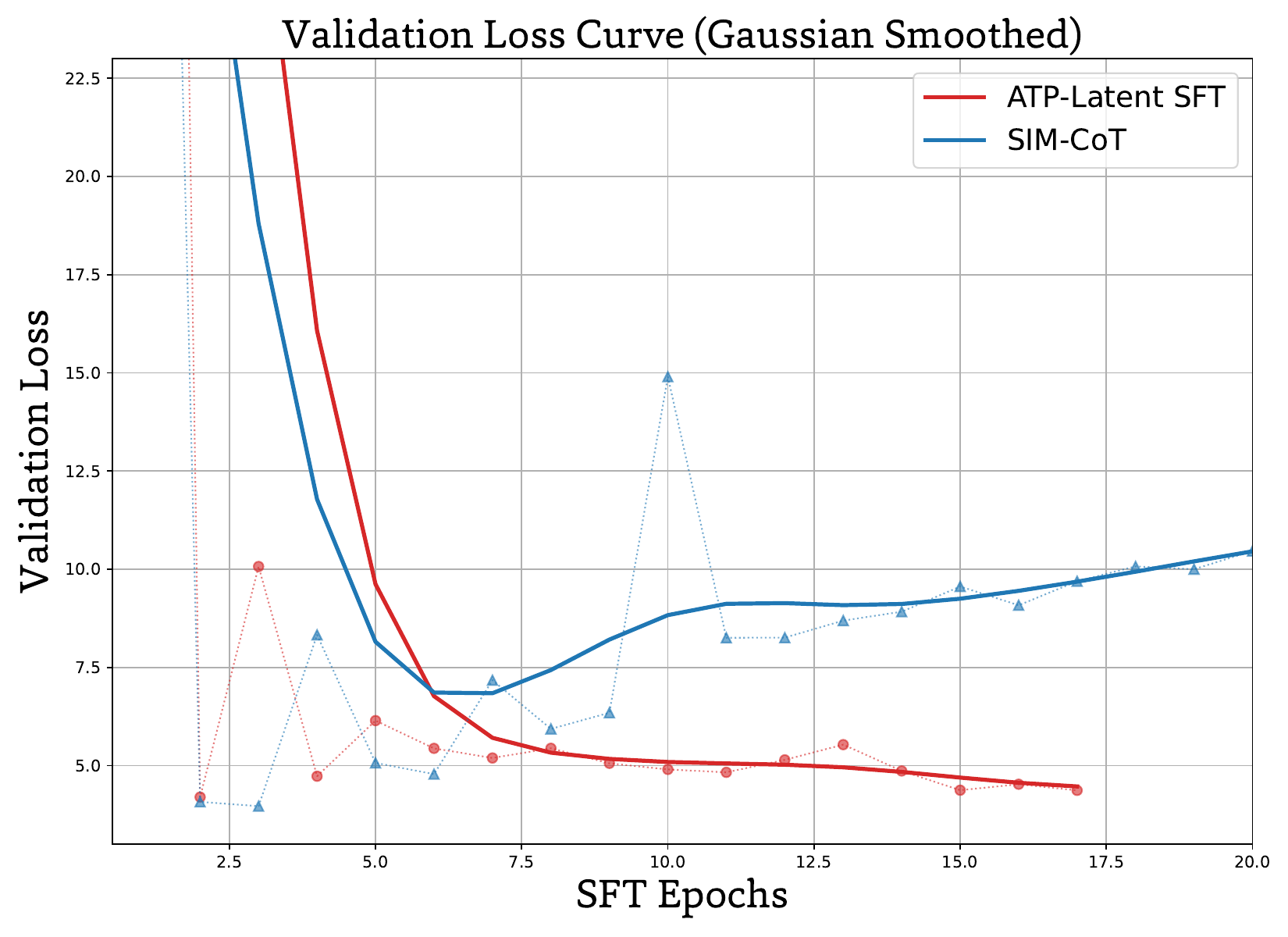}}
    \subfigure[Correctness vs Coherence Reward]{\includegraphics[width = 0.33\textwidth]{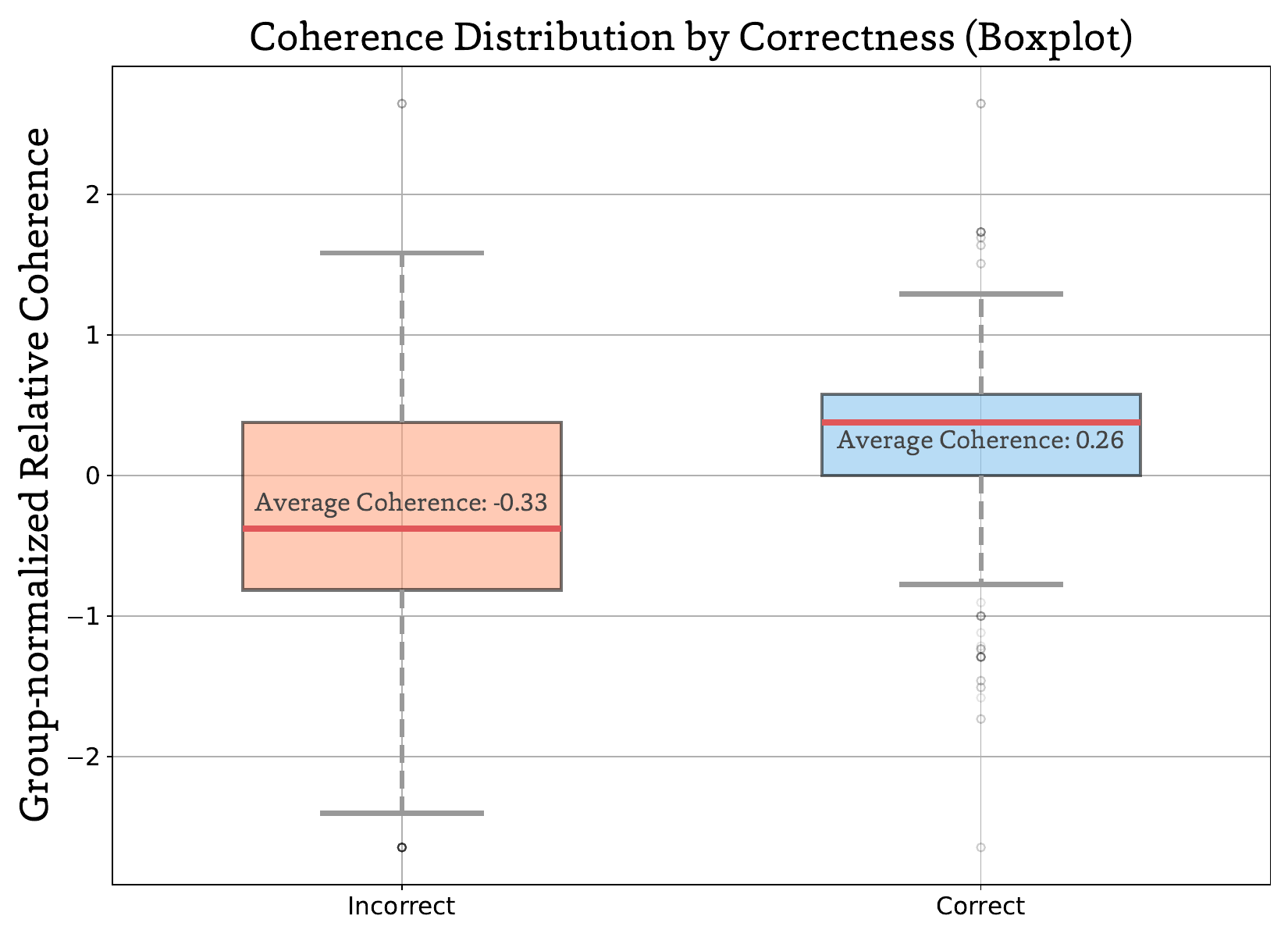}}
    \caption{Ablation study and case studies on ATP-Latent. (a) exhibits the validation accuracy curve during RL on GSM8K. (b) shows the loss curve during the SFT stage on the validation set of GSM8K-Aug (500 instances). (c) presents the relationship between the correctness of questions on the training dataset (RL split) and the coherence reward before the RL training in ATP-Latent.}\label{fig:abl}\vspace{-5pt}
\end{figure*}

\section{Experiment}
In this section, we evaluate the proposed ATP-Latent based on the \textit{LLaMA-3.2-1B-Instruct} LLM. 

\paragraph{Training \& Testing Settings} Following previous work \citep{hao2024training,wei2025sim,tan2025think}, we adopt the \textbf{GSM8K-Aug} \citep{deng2024explicit} with 385k instances as the training dataset. As shown in Figure \ref{fig:figure1} and \ref{fig:figure2}, the language CoTs in the dataset are built with equation steps. In training latent reasoning methods with RL stages, we divide the training data into two parts, 80\% for SFT and 20\% for RL. Detailed parameters are shown in Appendix \ref{experiment-detail}.

We follow \citep{wei2025sim} in adopting the GSM8K test set for in-domain, GSM-Hard \cite{gao2023pal}, SVAMP \citep{patel2021nlp}, and Multi-Arith \citep{roy2015solving} for out-of-domain evaluation.

\paragraph{Baselines} We adopt \textbf{latent reasoning methods} Coconut \citep{hao2024training}, SIM-CoT \cite{wei2025sim} (building onoconut), iCoT \cite{deng2024explicit}, and CoLaR \cite{tan2025think} as baselines. Besides, we introduce CoT-SFT for SFT with \textbf{language CoT steps} and Answer-SFT for supervising only the answer. All experiments are implemented on a node of 8× NVIDIA H200 GPUs (141 GB VRAM each). In obtaining baselines in a similar number of tokens, we set the compression factor in CoLaR to 5, $c=2$ for three curriculum stages in Coconut, and SIM-CoT. Please refer to Appendix \ref{baseline} for the detailed implementations of baselines.

\subsection{Results} 

Table~\ref{main} shows the performance of our proposed ATP-Latent and representative latent-reasoning baselines on four math reasoning benchmarks. We report answer accuracy (Acc; $\times$100 for clarity, higher is better) and the average number of generated tokens (\#Token; lower is better) to jointly measure correctness and generation efficiency. As shown, ATP-Latent achieves the strongest overall trade-off, reaching 47.7\% average accuracy with only 8.4 tokens on average, while being consistently competitive across all datasets; notably, it attains the outstanding 94.4\% on MultiArith. Compared to SIM-CoT and Coconut baselines, where we report both the re-implementation results and results reported in \citet{wei2025sim} (noted with "*"), ATP-Latent can demonstrate a +4.1\% accuracy and -3.3\% tokens compared to the reimplemented SIM-CoT result, demonstrating the benefits gained from doing active planning within a well-defined and smooth latent token representation space. In Appendix \ref{experiment-appendix}, we conduct more comparisons to baselines boosted with RL, where ATP-Latent can still demonstrate superiority.

\subsection{Main Ablations}
In Table \ref{main}, we also present ablation studies on the VAE (\textit{w/o} VAE for removing the $\boldsymbol{\sigma}_t$ prediction and KL loss, running the SIM-CoT-like method with stop-head over Coconut), the stop head (\textit{w/o} stop head for dropping the stop head, keeping the max stages to 3), and the RL stage (\textit{w/o} RL for the ATP-Latent model after the SFT stage). The ablation results validate the effectiveness of each proposed component: removing the VAE or the Stop Head leads to consistent accuracy drops (average -0.5\% and -0.8\%, respectively), and removing the RL stage yields the largest degradation (average -1.9\%), highlighting the importance of RL for improving latent reasoning. Furthermore, \textit{w} Only Coherence as Reward RL represents the unsupervised RL method shown in Section \ref{coh-only}, where using only the unsupervised coherence reward still improves the SFT-only model (average +0.9\%), indicating that coherence provides a useful unsupervised training signal.

\section{Discussion}

In this section, we conduct more ablation studies on components, as well as case studies over the interpretability and diversity of latent planning.

\subsection{More Ablation Studies}

\paragraph{Ablations on the Reward Function} In Table \ref{main}, we provide ablations over some components of ATP-Latent (i.e., VAE, Stop Head, and RL stage). In this subsection, we discuss the importance of introducing the coherence reward $R_\text{Coh}$ in the final reward $f(L,A)$. As shown in Figure \ref{fig:abl} (a), removing the coherence reward $R_\text{Coh}$ (the variant ATP-Latent-rl (without the coherence reward)) can clearly demonstrate inferior in-domain performance compared to the original ATP-Latent. It shows the significance of the coherence reward in RL training; we believe it can provide a good reweighting over positive samples, helping to find latent reasoning policies that conform to optimal properties.

\paragraph{Ablations on the VAE Training} As shown in Figure \ref{fig:abl}(b), by modeling the imitation process as training a VAE with a stop-head, ATP-Latent can demonstrate smoother convergence on the validation loss function in SFT compared to the SIM-CoT baseline (using the Coconut backbone).

\subsection{Case Study: Interpretability for Planning}
In this section, we discuss the validity of using coherence as an unsupervised reward and examine the changes in the VAE-decoded content to illustrate cases of the latent reasoning policy after the RL stage of ATP-Latent. 

\paragraph{The Validity of Coherence in Reward} As shown in Figure \ref{fig:abl} (a), using coherence as the only unsupervised reward can improve GSM8K performance in the first 800 steps. We believe that this benefits from the fact that the coherence can be a good unsupervised reward model for accuracy. As shown in Figure \ref{fig:abl} (c), we calculate the coherence on latent tokens generated by ATP-Latent after the SFT stage. The average group-normalized coherence value is 0.26 for correct examples and -0.33 for incorrect ones, demonstrating clear differences.

\begin{figure}[htbp]
    \centering
    \includegraphics[width=\linewidth]{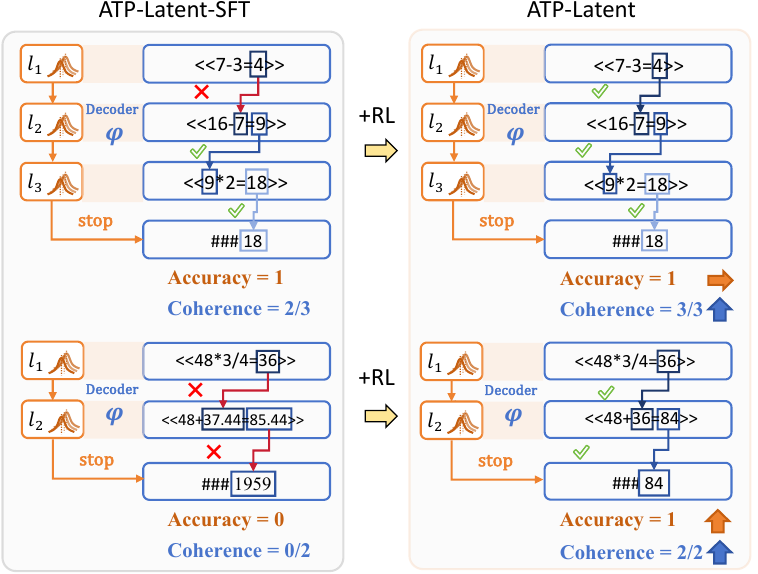}
    \caption{Examples of interpretability of latent tokens and improvements after RL. The accuracy and coherence can be improved.}
    \label{fig:inter}\vspace{-10pt}
\end{figure}

\paragraph{Examples for the Latent Reasoning Policy}

Similar to \citet{wei2025sim}, ATP-Latent can provide optional interpretability of latent tokens, using the VAE decoder $\phi$. As shown in Figure \ref{fig:inter}, the RL process can bring more consistent steps, as well as more accurate results. The improvement in consistency can be attributed to the introduction of $R_\text{Coh}$. In the second example, LLMs will prioritize correct and more coherent latent reasoning policies, proceeding with the planning process step by step, improving the probability of finding general and superior latent reasoning policies.

\begin{figure}[htbp]
    \centering
    \includegraphics[width=\linewidth]{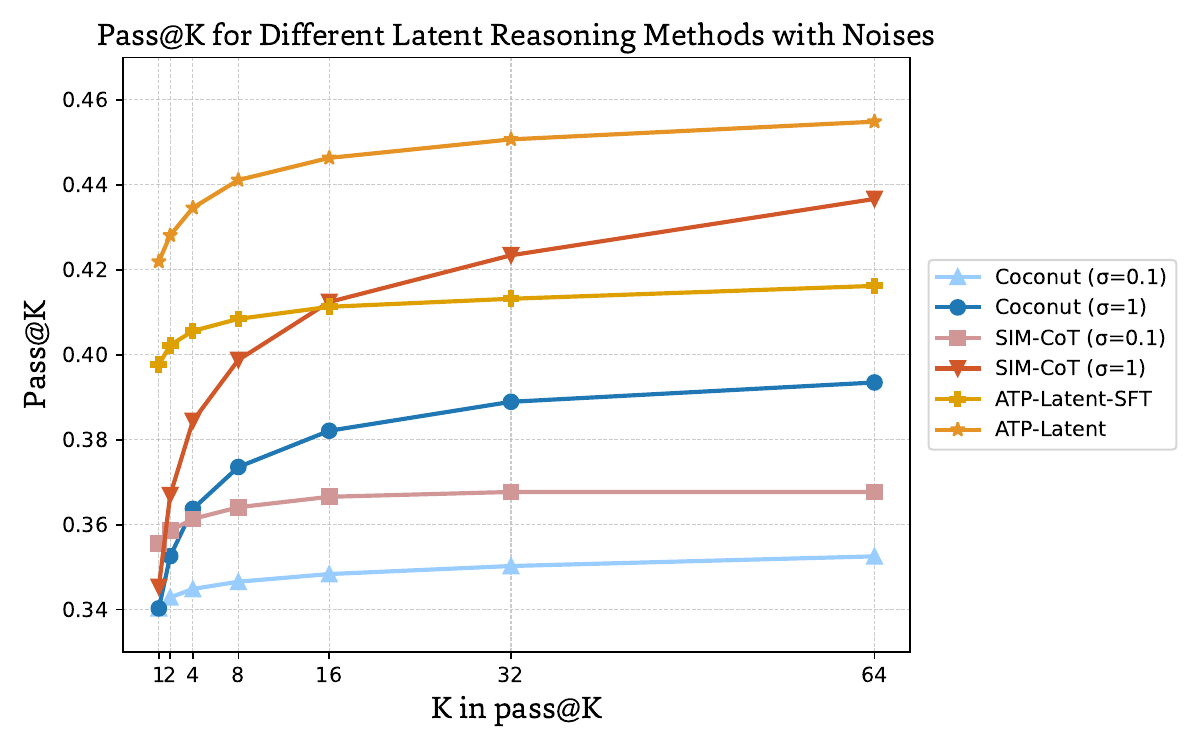}
    \caption{Pass@K curve for ATP-Latent and baselines. We run methods 64 times with Gaussian noise and different variances for Pass@K. The variances in ATP-Latent are predicted by itself.}
    \label{fig:passk}\vspace{-10pt}
\end{figure}

\subsection{Incentivize Planning Capacity Beyond Supervision} \label{incentivize}

To investigate whether the RL stage of ATP-Latent can incentivize the latent planning ability, we sample 64 runs with different $\boldsymbol{L}$ for Pass@K \citep{yue2025does}. Pass@K judges whether the correct answer occurs in every K runs. As shown in Figure \ref{fig:passk}, after the RL training process, compared to baselines and the ATP-Latent-SFT, ATP-Latent can show significantly higher Pass@K from K=1 to 64. It can reflect that ATP-Latent can develop latent planning capacities in the RL stage (examples are given in Appendix \ref{branchs}).

\section{Conclusion}
In this work, we highlight the significance of active planning over the latent token representation space and introduce ATP-Latent for actively planning in a well-defined and smooth latent space. Unlike prior latent reasoning methods that rely heavily on imitating single annotated CoT trajectories, ATP-Latent leverages a VAE framework and introduces a stop-generation mechanism to ensure a smoother and more expressive latent space. Moreover, by incorporating both the accuracy and coherence of VAE-decoded latent CoTs as rewards, our method provides soft constraints on the RL state space for more generalizable latent reasoning policies. Evaluations on four math reasoning benchmarks demonstrate that ATP-Latent achieves higher accuracy with significantly fewer generated tokens compared to existing baselines. ATP-Latent provides a promising direction for latent reasoning, and we will implement it on more tasks (e.g., natural language and complex LaTeX) in the future.

\section*{Impact Statement}

This paper presents work whose goal is to advance the field of machine learning. There are many potential societal consequences of our work, none of which we feel must be specifically highlighted here.

\bibliography{example_paper}
\bibliographystyle{icml2025}

%%%%%%%%%%%%%%%%%%%%%%%%%%%%%%%%%%%%%%%%%%%%%%%%%%%%%%%%%%%%%%%%%%%%%%%%%%%%%%%
%%%%%%%%%%%%%%%%%%%%%%%%%%%%%%%%%%%%%%%%%%%%%%%%%%%%%%%%%%%%%%%%%%%%%%%%%%%%%%%
% APPENDIX
%%%%%%%%%%%%%%%%%%%%%%%%%%%%%%%%%%%%%%%%%%%%%%%%%%%%%%%%%%%%%%%%%%%%%%%%%%%%%%%
%%%%%%%%%%%%%%%%%%%%%%%%%%%%%%%%%%%%%%%%%%%%%%%%%%%%%%%%%%%%%%%%%%%%%%%%%%%%%%%
\newpage
\appendix
\onecolumn

\section{Related Work}

\subsection{Latent Reasoning}
Latent reasoning methods decouple the reasoning process from explicit natural language and perform inference in the hidden space of the model. Existing latent methods can be generally divided into two categories: auto-aggressive methods and auxiliary strategies \cite{chen2025latentcot,zhu2025survey}. Token-wise auto-aggressive methods transform the conventional discrete-token CoT reasoning process into generating reasoning chains with dense latent embeddings \cite{hao2024training} or specialized tokens (e.g., pause \cite{goyal2024pause,zelikman2024quietstar}) one-by-one. These methods focus on transferring the original reasoning policy in the language domain to a latent embedding space, including curriculum learning (e.g., Coconut \cite{hao2024training}, LightThinker \cite{zhang2025lightthinker}, self-distillation (e.g., CODI \cite{shen2025codi}), and one-shot compression (e.g., CoLaR \cite{tan2025think}, SynAdapt \cite{wang2025synadapt}). Due to the existing token-wise auto-aggressive methods completely treating the language CoTs as the label, these methods can hardly surpass or even reach the performance level of discrete-token CoT reasoning. So empirically, these methods can effectively reduce the token cost compared to language CoT, but clear performance drop.

Latent reasoning methods with auxiliary strategies (e.g., SoftCoT \cite{xu2025softcot}) generate latent embeddings from an auxiliary module and inject them into the main model for the following discrete-token CoT \cite{cheng2024compressed,xu2025softcot++,kong2025latent,liu2025marcos}. Unlike auto-regressive latent reasoning methods, auxiliary strategies can effectively improve the performance of the original LLM, but at the cost of running efficiency.

This paper focuses on designing a token-wise auto-aggressive latent reasoning method. The proposed ATP-Latent highlights the fact that the optimal latent reasoning policy can only be obtained by imitation, and first builds the step-by-step imitation process as a VAE.

\subsection{Latent Reasoning with Reinforcement Learning}

As briefly illustrated in Section \ref{section3}, there are several existing works or contemporary on using RL for latent reasoning \citep{tan2025think,ozeren2025reinforcement,ning2025learning}. These methods employ Gaussian noise and reparameterization techniques, addressing the issues of latent tokens often being deterministic, lacking randomness, and the gradient problem for RL. These methods often rely on searching for the optimal latent reasoning policy within a given Gaussian space using batch data. In contrast, ATP-Latent emphasizes using coherence to first orthogonalize a defined legitimate latent planning approach, avoiding shortcuts \citep{zhang2025latent,shen2025codi} that directly answer without considering intermediate latent reasoning steps.

Besides RL methods, some other methods incorporate noise for planning, \citet{you2025parallel} propose to do test-time compute, using a fine-tuned verifier for majority voting. Although accuracy improvements are made, this method will harm the effective nature of latent reasoning. 

There are also other reinforcement learning methods for latent reasoning methods with auxiliary strategies. \citet{yue2025hybrid} fine-tunes LLMs to pass hidden states between tokens to enrich the LLM's discrete-token reasoning.

\subsection{Soft-Thinking Reasoning}
As a special category of reasoning method, soft-thinking reasoning passes the summation of token embeddings weighted by output probabilities as the LLM step. It mainly aims at conveying abstract concepts that cannot be represented by one language token and improving information density \citep{zhang2025soft,wu2025llms,zhuang2025text,butt2025soft,zheng2025soft}. Therefore, although soft-thinking can demonstrate better performance than discrete-token in many cases, and the RL method SofT-GRPO based on the soft-thinking reasoning pattern can show better Pass@K performance than discrete-token GRPO, such methods cannot lead to the significant efficiency improvements seen in latent-reasoning.

\subsection{Latent Reasoning in Vision Language Model}

Recent work also tries to adopt the latent reasoning idea in visual tasks \citep{li2025latent,wang2025monet,pham2025multimodal,chen2025reasoning,ma2025cocova,qin2025chain,yu2025vismem,dong2025interleaved}, and some of them adopt the RL idea in \citet{tan2025think}. But as far as we acknowledge, none of them adopts the same coherence idea in the proposed ATP-Latent.

\newpage
\section{Prompt for ATP-Latent} \label{prompt}
In this section, we will introduce the format for ATP-Latent as well as the settings for the format reward $R_\text{Format}$. The prompt for latent reasoning contains only the question as follows:

\begin{dialogbox}[Prompt for ATP-Latent.]\label{latent prompt}
\textcolor{Gray}{user:} \textcolor{brown}{Janet’s ducks lay 16 eggs per day. She eats three for breakfast every morning and bakes muffins for her friends every day with four. She sells the remainder at the farmers' market daily for \$2 per fresh duck egg. How much in dollars does she make every day at the farmers' market?} Let's think step by step and output the final answer after \#\#\# \textcolor{Red}{$<$latent$>$}

~\\
\textcolor{Gray}{assistant:} \textcolor{Orange}{\ldots Latent tokens \ldots}\textcolor{Red}{$</$latent$>$}\#\#\# 18

\end{dialogbox}

We set $R_\text{Format}$ as 1 if the \#\#\# can be successfully output after \textcolor{Red}{$<$latent$>$} and 0 elsewise. The decoder $\phi$ inputs only the $c$ latent tokens in one step for the decoded content.

In Table \ref{main}, the prompts of CoT-SFT and Answer-SFT baselines are as follows.

\begin{dialogbox}[Prompt for CoT-SFT in Tabel \ref{main}]
\textcolor{Gray}{user:} \textcolor{brown}{Janet’s ducks lay 16 eggs per day. She eats three for breakfast every morning and bakes muffins for her friends every day with four. She sells the remainder at the farmers' market daily for \$2 per fresh duck egg. How much in dollars does she make every day at the farmers' market?} Let's think step by step and output the final answer after \#\#\#

~\\
\textcolor{Gray}{assistant:} \textcolor{Orange}{\ldots Language CoT \ldots}\#\#\# 18

\end{dialogbox}

\begin{dialogbox}[Prompt for Answer-SFT in Tabel \ref{main}]
\textcolor{Gray}{user:} \textcolor{brown}{Janet’s ducks lay 16 eggs per day. She eats three for breakfast every morning and bakes muffins for her friends every day with four. She sells the remainder at the farmers' market daily for \$2 per fresh duck egg. How much in dollars does she make every day at the farmers' market?} Output the final answer after \#\#\#

~\\
\textcolor{Gray}{assistant:} \#\#\# 18

\end{dialogbox}

In counting the number of generated tokens of all methods, we consider all the tokens generated by LLMs (i.e., except \textcolor{Red}{$<$latent$>$}, \textcolor{Red}{$</$ latent$>$}).

\newpage
\section{Detailed parameters} \label{experiment-detail}

The detailed parameters of ATP-Latent are listed as follows. We use the \textit{LLaMA-3.2-1B-Instruct} LLM for both the reasoning LLM $\theta$ and decoder $\phi$. Both the stop head and the latent head are two-layer MLPs using the GELU \citep{hendrycks2016gaussian} activation.

\begin{table}[H]
    \centering
    \renewcommand\arraystretch{1}
    \caption{Parameters of ATP-Latent}
    \begin{tabularx}{0.8\textwidth}{
    l@{\hskip 0.8in}
    X
}
    \toprule
 Parameter&  Value\\
 \midrule
 \multicolumn{2}{c}{\textsc{ATP-Latent} SFT stage} \\
\midrule
Optimizer & AdamW\\
Batch size & 256\\
Learning rate & 1e-4\\
KL loss factor $\beta$ (KL normalized with $d$) & 1e-3\\
Weight decay & 0.01\\
Maximum stages & 10\\
Number of epochs per stage & 2\\
Number of latent tokens for each stage $c$ &2\\
Number of epochs & 15\\
GPU usage & 8\\
Data percentage & 80\%\\
 \midrule
 \multicolumn{2}{c}{\textsc{ATP-Latent} RL stage} \\
\midrule
Optimizer & AdamW\\
Batch size & 16\\
Learning rate & 1e-6\\
Maximum answer length & $16$ tokens \\
Maximum stages & 10\\
Number of latent tokens for each stage $c$ &2\\
Sampling temperature & 1.0 \\
(top-p, top-k) & (1, -1) \\
Number of epochs & 1\\
Clipping threshold $\epsilon$ & 0.2\\
KL loss factor in GRPO \citep{shao2024deepseekmath} & 0\\
GPU usage & 1\\
Data percentage & 20\%\\
 \midrule
 \multicolumn{2}{c}{\textsc{ATP-Latent} Testing} \\
\midrule
Batch size & 1\\
Maximum answer length & $16$ tokens \\
Maximum stages & 10\\
Number of latent tokens for each stage $c$ &2\\
Sampling temperature & 0 \\
(top-p, top-k) & (1, -1) \\
 \bottomrule
    \end{tabularx}
    \label{tab:nf}
\end{table}
 
\newpage
\section{Supplementary Experiments} \label{experiment-appendix}

\subsection{Comparison with Baselines + RL}

Table \ref{main} demonstrates the performance of ATP-Latent. In this part, we compare ATP-Latent with more latent reasoning methods and the reinforcement learning (RL) fine-tuning. Metrics include accuracy (Acc) and average reasoning length (\#Token).

Results in Table \ref{table2} show that RL fine-tuning brings little improvement for CoLaR, but leads to substantial performance gains for Coconut, Coconut + SIM-CoT (fine-tuning for SIM-CoT over a pre-trained Coconut, the one shown in Figure \ref{fig:simcot}), and especially the proposed ATP-Latent method. After RL tuning, ATP-Latent achieves the highest overall average accuracy (47.7\%), outperforming all baselines. The RL process of ATP-Latent demonstrates more significant improvement compared to the Coconut + SIM-CoT baseline, indicating the VAE, stop-head, and coherence reward can provide better state space for the RL improvement.

\begin{table}[H]
\centering
\caption{
Comparison of latent reasoning methods with the following RL fine-tuning. We consider CoLaR, Coconut, Coconut + SIM-CoT (fine-tuning SIM-CoT over pretrained Coconut), and the proposed ATP-Latent.
}
\renewcommand\arraystretch{1.1}
\setlength{\tabcolsep}{2mm}
\resizebox{\textwidth}{!}{
\begin{tabular}{l|cc|cccccc|cc}
\bottomrule[0.5mm]
     & \multicolumn{2}{c|}{GSM8K} & \multicolumn{2}{c}{GAM-hard} & \multicolumn{2}{c}{MultiArith} & \multicolumn{2}{c|}{SVAMP} & \multicolumn{2}{c}{Avg} \\ \hline
\multicolumn{1}{c|}{Methods} & Acc& \#Token & Acc & \#Token& Acc  & \#Token & Acc& \#Token & Acc   & \#Token \\ \hline \hline
CoLaR \cite{tan2025think}& 25.7 & 10.0  & 5.7 & 11.5   & 86.8 & 7.2     & 49.9 & 7.1   & 42.0  & 8.9     \\
+RL  & 25.5 & 10.0  & 5.9 & 11.4   & 86.8 & 7.2     & 49.0 & 7.1   & 41.8{\small\textcolor{Red}{(-0.2)}}  & 8.9     \\ \hline
Coconut \cite{hao2024training}& 35.6 & 9.2   & 8.2 & 9.8    & 86.1 & 9.0     & 37.0 & 9.1   & 41.7  & 9.3     \\
+RL  & 40.0 & 9.2   & 9.3 & 9.8    & 90.0 & 9.0     & 42.7 & 9.0   & 45.5{\small\textcolor{Green}{(+3.8)}}  & 9.3     \\ \hline
Coconut + SIM-CoT \cite{wei2025sim}   & 40.0 & 9.2   & 9.7 & 9.8    & 88.9 & 9.0     & 45.3 & 9.0   & 46.0  & 9.3     \\
+RL  & 42.1 & 9.2   & 10.2& 9.8    & 90.6 & 9.0     & 45.7 & 9.0   & 47.1{\small\textcolor{Green}{(+1.1)}} & 9.3     \\ \hline
ATP-Latent-SFT (Ours) & 40.0 & 9.5   & 9.3 & 9.7    & 90.0 & 7.1     & 43.7 & 6.5   & 45.8  & 8.2     \\
+RL (Ours)  & 42.3 & 9.8   & 9.8 & 10.0   & 94.4 & 7.1     & 44.2 & 6.8   & \textbf{47.7}{\small\textcolor{Green}{(+1.9)}} & 8.4     \\ \toprule[0.5mm]
\end{tabular}
}
\label{table2}
\end{table}

\subsection{Comparison on the Self-Extending Ability}

As mentioned in \citet{wei2025sim}, the Coconut has a latent instability issue. Coconut pre-trained on a specific maximum stage (3 in their and our paper) cannot extend to larger stages, and SIM-CoT can mitigate such degradation in performance when generalizing to larger stages (i.e., larger maximum numbers of latent tokens). 

ATP-Latent can totally solve such an instability issue by introducing a stop mechanism. As shown in Table \ref{table3}, ATP-Latent and its SFT model will not drop performance when the number of maximum stages (i.e., the maximum number of latent tokens) improves.

\begin{table}[H]
\centering
\caption{
The performance on GSM8K of Coconut, SIM-CoT, ATP-Latent, and ATP-Latent-SFT with different numbers of latent stages (i.e., number of latent tokens).
}
\renewcommand\arraystretch{1.1}
\setlength{\tabcolsep}{3.5mm}
\resizebox{\textwidth}{!}{
\begin{tabular}{l|c|ccccccccc}
\bottomrule[0.5mm]
\multicolumn{1}{l|}{}   & \#Stage & 1    & 2    & 3    & 4    & 5    & 7    & 10   & 15   & 20   \\ \hline\hline
\multirow{2}{*}{Coconut}& Acc     & 25.2 & 33.1 & \textbf{35.6} & 34.7 & 33.8 & 30.7 & 26.5 & 21.0 & 12.7 \\
& \#Token & 5.2  & 7.2  & 9.2  & 11.2 & 13.2 & 17.2 & 23.2 & 33.2 & 43.2 \\ \hline
\multirow{2}{*}{SIM-CoT}& Acc     & 19.6 & 29.0 & \textbf{35.9} & 34.4 & 31.3 & 28.1 & 25.3 & 18.3 & 15.1 \\
& \#Token & 5.2  & 7.2  & 9.2  & 11.2 & 13.2 & 17.2 & 23.2 & 33.2 & 43.3 \\ \hline
\multirow{2}{*}{ATP-Latent-SFT} & Acc     & 12.4 & 25.1 & 34.0 & 39.1 & 39.7 & 39.8 & 40.0 & \textbf{40.3} & 40.2 \\
& \#Token & 5.2  & 7.1  & 8.4  & 9.1  & 9.4  & 9.5  & 9.5  & 9.5  & 9.5  \\ \hline
\multirow{2}{*}{ATP-Latent}     & Acc     & 13.8 & 26.4 & 36.4 & 40.8 & 42.2 & 42.4 & 42.3 & \textbf{42.5} & 42.4 \\
& \#Token & 5.1  & 7.1  & 8.5  & 9.3  & 9.7  & 9.8  & 9.8  & 9.8  & 9.9  \\ \toprule[0.5mm]
\end{tabular}
}
\label{table3}
\end{table}

\newpage
\subsection{Case Study: Exploring Different Ways} \label{branchs}

Section \ref{incentivize} shows that the RL process of ATP-Latent-SFT can actually develop new patterns, developing diverse reasoning paths, thus improving Pass@K \cite{yue2025does,hu2026rewarding}. In Figure \ref{fig:inter}, we decode the latent tokens to language steps and show the tree of the probability of going into language tokens representing different language steps. It directly exhibits that after RL training, ATP-Latent is able to generate a wider variety of valid reasoning paths (“branches”) for the same problem, rather than always generating latent paths representing one language label. This demonstrates that ATP-Latent’s RL stage encourages more generalizable and flexible reasoning by promoting exploration of multiple valid solutions.
\begin{figure}[H]
    \centering
    \includegraphics[width=0.65\linewidth]{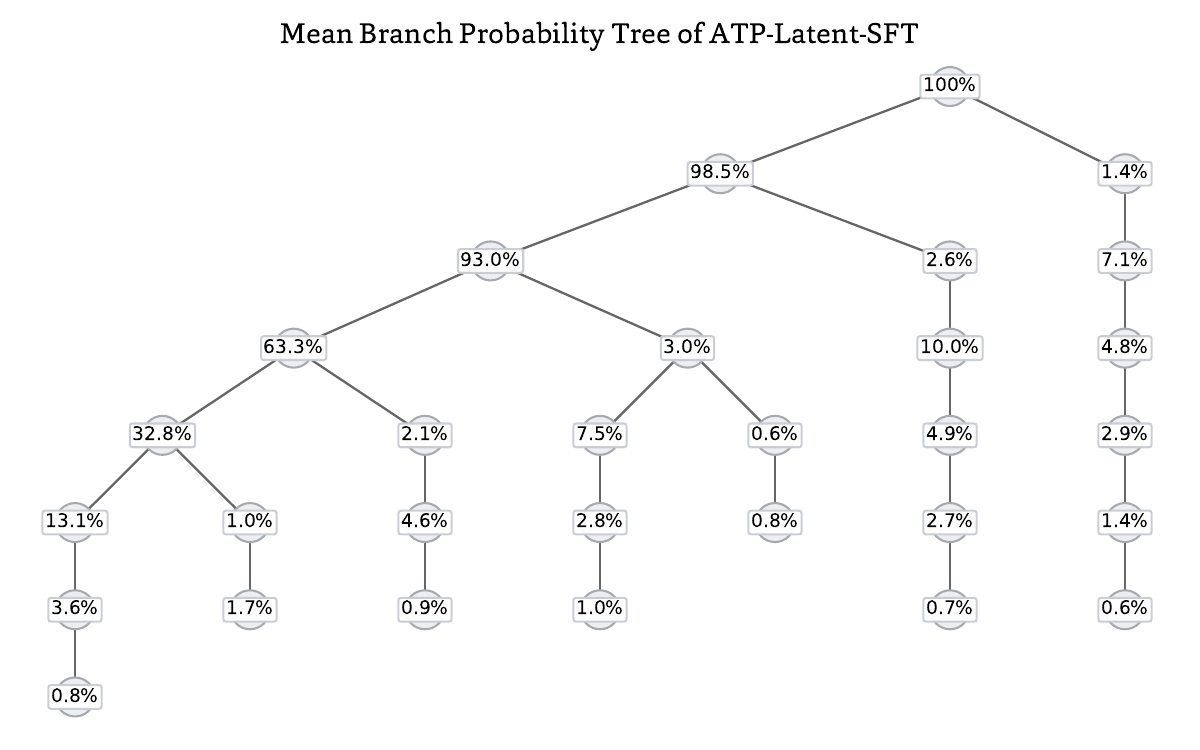}
    \includegraphics[width=0.7\linewidth]{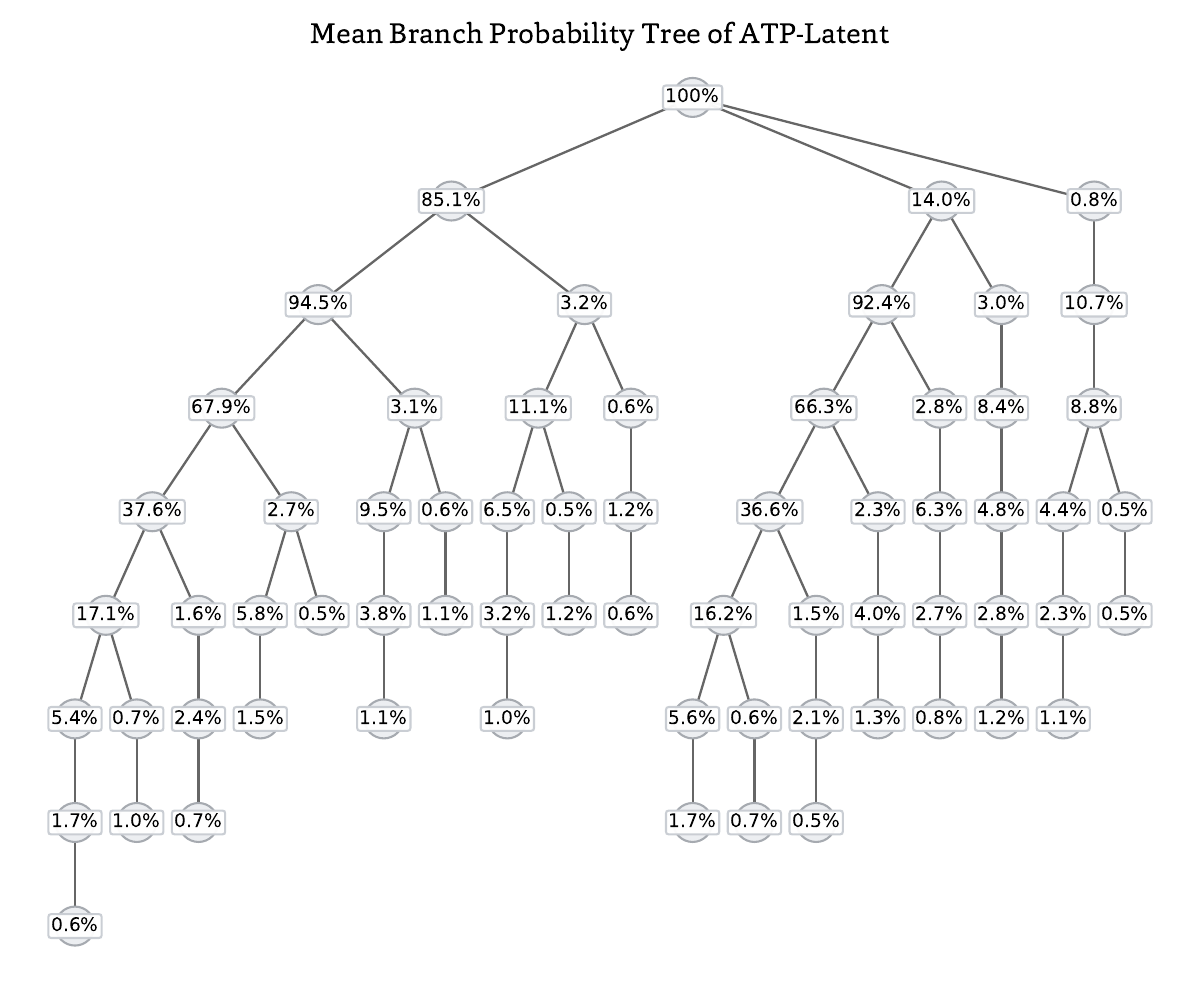}
    \caption{Pattern change after the RL training of ATP-Latent. The upper is for ATP-Latent-SFT, and the lower shows the pattern of ATP-Latent. We run each instance 64 times, decode each latent token to language steps by decoder, and count the probability of going into the top-1 branch or the branches with lower probabilities. Numbers in branches represent the probability of going into that branch. We do not show the branches with probability $<1\%$ and the branches stopped by the stop head, so the summation of probabilities may not be 100\%.}
    \label{fig:inter}
\end{figure}

\newpage
\section{Baselines \& Datasets \& Licenses} \label{baseline}

\subsection{Baselines}

Based on the \textit{\textbf{LLaMA-3.2-1B-Instruct}} LLM, we include SFT baselines for CoT-SFT, Answer-SFT. We also include latent reasoning baselines, CoLaR \citep{tan2025think}, iCoT \citep{deng2023icot}, Coconut \cite{hao2024training}, and SIM-CoT \citep{wei2025sim}

\paragraph{SFT Baselines} We implement SFT baselines based on the LLaMA-Factory \citep{zheng2024llamafactory} framework. The prompts of CoT-SFT and Answer-SFT are given in Appendix \ref{prompt}.

\paragraph{Latent Reasoning Baselines} There may be complex training stages (e.g., SFT warm-up) in reproducing baselines (e.g., SIM-CoT); we fine-tune all latent reasoning from the base LLM using their given code. The reproduction results shown in Table \ref{main} can generally match their report with "*".

\subsection{Datasets}

We follow \citep{wei2025sim} in adopting the GSM8K test set for in-domain, GSM-Hard \cite{gao2023pal}, SVAMP \citep{patel2021nlp}, and Multi-Arith \citep{roy2015solving} for out-of-domain evaluation.

\subsection{Licenses}

For Base LLM, Dataset, and frameworks, we list their Licenses in Table \ref{license}.

\begin{table}[H]
\centering
\setlength{\tabcolsep}{1mm}
\renewcommand\arraystretch{1.2}
\caption{A summary of licenses.}
\resizebox{\textwidth}{!}{
\begin{tabular}{llll}
\bottomrule[0.8mm]
Resources & Type    & License& URL    \\ \midrule[0.3mm]
LLaMA-3.2-1B-Instruct & Base LLM    &  Llama 3.2 License &  \url{https://huggingface.co/meta-llama/Llama-3.2-1B-Instruct} \\ \midrule[0.3mm]
LLaMA-Factory & SFT-framework    &  Apache-2.0 license &  \url{https://github.com/volcengine/verl} \\ 
CoLaR & Baseline    &  Apache-2.0 license &  \url{https://github.com/xiaomi-research/cola} \\
Coconut & Baseline    &  MIT License &  \url{https://github.com/facebookresearch/coconut} \\
SIM-CoT & Baseline    &  Apache-2.0 license &  \url{https://github.com/InternLM/SIM-CoT} \\ \midrule[0.3mm]
GSM8K, GSM-Hard, SVAMP & Dataset    & MIT License    &\url{https://github.com/xiaomi-research/colar} \\
Multi-Arith & Dataset    & Available Online    &\url{https://github.com/xiaomi-research/colar} \\ 
GSM8K-Aug & Dataset    & Available Online    &\url{https://huggingface.co/datasets/whynlp/gsm8k-aug} \\ \toprule[0.8mm]
\end{tabular}}
\label{license}
\end{table}

%%%%%%%%%%%%%%%%%%%%%%%%%%%%%%%%%%%%%%%%%%%%%%%%%%%%%%%%%%%%%%%%%%%%%%%%%%%%%%%
%%%%%%%%%%%%%%%%%%%%%%%%%%%%%%%%%%%%%%%%%%%%%%%%%%%%%%%%%%%%%%%%%%%%%%%%%%%%%%%

\end{document}